# Title

Position-wise optimizer: A nature-inspired optimization algorithm

# Author

Amir Valizadeh, M.D., Neuroscience Institute, Tehran University of Medical Sciences, ORCID
Keshavarz Blvd., Tehran, Iran
Phone: 0982164053334; Email: thisisamirv@gmail.com

# Abstract

The human nervous system utilizes synaptic plasticity to solve optimization problems. Previous studies have tried to add the plasticity factor to the training process of artificial neural networks, but most of those models require complex external control over the network or complex novel rules. In this manuscript, a novel nature-inspired optimization algorithm is introduced that imitates biological neural plasticity. Furthermore, the model is tested on three datasets and the results are compared with gradient descent optimization.

# Introduction

## Neural plasticity in biological neural networks

"Neural plasticity" refers to the capacity of the nervous system to modify itself, functionally and structurally, in response to experience or injury[1]. The concept of neural plasticity is mostly built upon the principle of Hebb's rule: if a neuron repeatedly takes part in making another neuron fire, the connection between them is strengthened ("neurons that fire together, wire together")[2]. Neural plasticity is possible through synaptic plasticity. Synaptic plasticity refers to the activity-dependent modification of the strength of synaptic transmission at preexisting synapses (i.e., change in synaptic weight)[3]. Synaptic plasticity has long been considered to have a central role in learning and memory formation[4].

While in the central nervous system plasticity is mainly based on adaptive changes in neural circuitries and synaptic reorganization, in the peripheral nervous system it is predominantly based on axonal re-growth and neuron addition[5], which is a much slower, restricted process. Considering this fact, it is proposed that the learning process and formation of new memories are mostly due to modifications in "deeper" (i.e., central) synapses, and the impact of synaptic plasticity of "shallower" (i.e., peripheral) synapses is less influential in those processes.

## Optimization problem

An optimization problem is a problem of finding the best solution from all feasible solutions[6]. Machine learning (ML) problems, once formulated, can be viewed as optimization problems. For supervised learning problems, the goal would be to find an optimal mapping function $f(x)$ to minimize the loss function ($L$) of the training samples[7],

$$\min_{\theta} \frac{1}{N} \sum_{i=1}^{N} L\left(y^i, f(x^i, \theta)\right),$$

(1)

where $N$ is the number of training samples, $\theta$ is the parameter of the mapping function, $x^i$ is the feature vector of the $i$th samples, and $y^i$ is the corresponding label.

The gradient descent method is the earliest and most common optimization method. The idea is to update parameters iteratively in the opposite direction of the gradients of the objective function[8]. The objective function could be written as,

$$L(\theta) = \frac{1}{2N} \sum_{i=1}^{N} \left(y^i - f_\theta(x^i)\right)^2,$$

(2)

where,

$$f_\theta(x) = \sum_{j=1}^{D} \theta_j x_j,$$

(3)

and $D$ is the number of input features. The gradient descent method alternates between the following two steps until it converges[7]:

1) Derive $L(\theta)$ for $\theta_j$ to get the gradient corresponding to each $\theta_j$:

$$\frac{\partial L(\theta)}{\partial \theta_j} = -\frac{1}{N} \sum_{i=1}^{N} \left(y^i - f_\theta(x^i)\right) x_j^i.$$

(4)

2) Update each $\theta_j$ in the opposite of the gradient direction to minimize the objective function:

$$\theta_j' = \theta_j + \eta \cdot \frac{1}{N} \sum_{i=1}^{N} \left(y^i - f_\theta(x^i)\right) x_j^i,$$

(5)

where $\eta$ is the learning rate.

Previous research has proposed that the human brain might also be using a gradient-based method to address optimization problems[9]. Yet again, there is a huge gap between human-level performance and machine performance in a variety of tasks, such as visual classification where a machine requires dozens of images of an object to be able to accurately recognize it in unseen examples, while a human agent probably needs just a couple of examples. Such astonishing learning rates raise the question of how could human brain's solution to optimization problems differs from that of machines (most notably, gradient descent)?

## Adding the plasticity factor to artificial neural networks

Artificial neural networks which employ backpropagation based on the gradient descent method have been used extensively in ML[10]. There have been previous studies that have tried to add the plasticity factor to the training process of artificial neural networks[11–13]. However, most of these models require complex external control over the network or complex novel rules, which in turn increase the computational burden of the model. In this manuscript, it is intended to implement synaptic plasticity in ANNs in a similar manner to that of the human nervous system. This model is supposed to be easier to implement, while also achieving faster learning rates.

To better implement such a model, a set of criteria were defined for the model to satisfy:

1. Local computation: An ANN performs computation only based on the inputs it receives from other neurons. This input is weighted by the strengths of its synaptic connections.
2. Global plasticity: The amount of change in synaptic weights is not only dependent on the activity of the two neurons the synapse connects, but also on the relative position of the ANN.
3. Minimal external control: The ANNs perform computations with as little external control.

# Methods

## Position-wise optimizer

The new model proposed in this research was named "position-wise optimizer" as its most prominent feature is to take the ANN's relative position to other ANNs into account for updating the parameters. Due to the chain rule, updating the weights of one layer requires computing the gradients of the weights of all the deeper layers[14]. In the standard gradient descent model, each parameter of the model (i.e., weights of the synapses, $\theta_j$) gets an update after each epoch (when the learning algorithm sees the complete dataset once). This process is not completely in line with the synaptic plasticity model in biological networks where synaptic weight change is more restricted for the peripheral synapses than the central synapses. To imitate the synaptic plasticity function of biological neural networks, the model must be altered for the deeper ANNs to be able to change their synaptic weight more frequently than the shallower ANNs.

The proposed model in this paper achieves that goal in the following steps:

I. Following a complete forward pass of the network, only the gradients for the last layer ($L$) will be computed. Afterward, the parameters (synaptic weights) of the last layer get an update. The reason behind this is to train the ANNs in the last layer to minimize the objective function based on input from shallower layers while holding them constant.

II. Another forward pass only using the last layer is done to be able to calculate the new gradients following the update in the previous step.

III. Gradients for the layer before the last layer ($L-1$) and the last layer ($L$) will be computed based on the new objective function value. Afterward, these two layers' parameters (synaptic weights) get an update. Again, the reason behind this is to train the ANNs in these layers to minimize the objective function based on input from shallower layers while holding them constant.

IV. The same process is repeated in a layer-wise manner up to the first layer of the network. In this process, deeper ANNs are trained to achieve a low objective function while the parameters of shallower ANNs (and thus their output to the deeper ANNs) change less frequently, compared to the times of change in the parameters of the deeper layers.

A simplified illustration of the proposed model is compared to the gradient descent model in **Figure 1**.

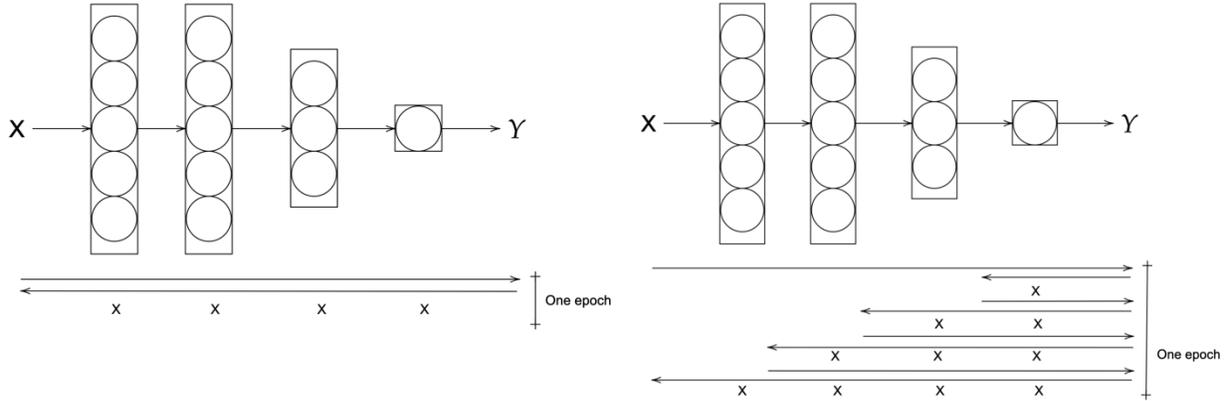

**Figure 1**; One epoch of the gradient descent optimizer (left) vs one epoch of the position-wise optimizer (right). "X" indicates weight update for the corresponding layer.

## Model implementation

The implementation of the model, compared with the gradient descent model, is presented in the following table:

| **Algorithm 1** Batch gradient descent | **Algorithm 2** Position-wise optimizer |
|---|---|
| **Require:** input $X$, weight $W$, output $Y$, desired output $D$, number of layers $L$, loss $E$, learning rate $\eta$ ||
| 1    **for** $l$ **in range** $(L-1)$ **do:** | 1    **for** $l$ **in range** $(L-1)$ **do:** |
| 2        $Y^{(l)} \leftarrow W^{(l)} X^{(l)}$ | 2        $Y^{(l)} \leftarrow W^{(l)} X^{(l)}$ |
| 3        $X^{(l+1)} \leftarrow \phi(Y^{(l)})$ | 3        $X^{(l+1)} \leftarrow \phi(Y^{(l)})$ |
| 4    Compute $E$ | 4    **for** $i$ **in range** $(L-1)$ **do:** |
| 5    $\Delta Y^{(L)} = Y^{(L)} - D$ | 5        **for** $l$ **in range** $(L-1, L-1-i)$ **do:** |
| 6    **for** $l$ **in range** $(L-1, 0)$ **do:** | 6           Compute $E$ |
| 7        $\Delta Y^{(l)} \leftarrow \Delta X^{(l+1)} \phi'(Y^{(l)})$ | 7           **if** $l = L-1$ **then:** |
| 8        $\Delta W^{(l)} \leftarrow \Delta Y^{(l)} X^{(l)^T}$ | 8              $\Delta Y^{(l)} = Y^{(l)} - D$ |
| 9        **if** $l \geq 1$ **then:** | 9           **else:** |
| 10           $\Delta X^{(l)} \leftarrow W^{(l)^T} \Delta Y^{(l)}$ | 10              $\Delta Y^{(l)} \leftarrow \Delta X^{(l+1)} \phi'(Y^{(l)})$ |
| 11    **for** $l$ **in range** $(L-1, 0)$ **do:** | 11           $\Delta W^{(l)} \leftarrow \Delta Y^{(l)} X^{(l)^T}$ |
| 12        $W^{(l)} \leftarrow W^{(l)} - \eta \Delta W^{(l)}$ | 12           **if** $l \geq 1$ **then:** |

| 13 | $\Delta X^{(l)} \leftarrow W^{(l)^T} \Delta Y^{(l)}$ |
|----|------------------------------------------------------|
| 14 | $W^{(l)} \leftarrow W^{(l)} - \eta \Delta W^{(l)}$ |
| 15 | $Y^{(l)} \leftarrow W^{(l)} X^{(l)}$ |
| 16 | $X^{(l+1)} \leftarrow \phi(Y^{(l)})$ |

## Experiments

The model was implemented in Python 3.9. The network consisted of 3 hidden layers of dimensions 20, 7, and 5 respectively. Weights were randomly initialized based on normal Xavier initialization:

$$w^{[l]} \sim \mathcal{N}(\mu, \sigma); where\ \sigma = \sqrt{\frac{2}{n^{[l]} + n^{[l-1]}}},$$

*(6)*

where $w^{[l]}$ is the weight matrix corresponding to layer $l$.

For binary classification tasks, the objective function was the cross-entropy loss, which is computed as follows:

$$E = -\frac{1}{N} \sum_{i=1}^{N} [y^i \log(D^i) + (1 - y^i) . \log(1 - D^i)].$$

*(7)*

For multi-class classification tasks, the objective function was the additive margin softmax loss. The bias term (b) for each layer was initially set to zero.

After initiating the weights, each weight was copied into two new variables, each used for one algorithm (gradient descent optimizer and position-wise optimizer). The network had 4 activation units. Starting from the activation unit of the first layer to the last, they included: ReLU, ReLU, ReLU, and Sigmoid activation functions. While training each model, at the end of each iteration, the loss for the model was computed and stored in a list data structure. Both models were trained using the same constant learning rate ($\eta$). Both algorithms were set to terminate the training process when the loss fell below a desirable value. After the termination of each training process, the time elapsed to train the model was measured using the "time" module. Finally, calculated losses were plotted against the number of epochs for each algorithm, using the "matplotlib" module.

# Results

## Experiment 1; In-house dataset for cat vs. non-cat classification

This experiment was a binary classification task. The dataset consisted of 209 images. Each image was of size: (64, 64, 3). At first, the data were imported into Python using the "h5py" module. Images were reshaped into one big vector, used for the training process. The input layer was a matrix with the dimensions of (12288, 209).
The training process results for both models are presented in **Figure 2**.

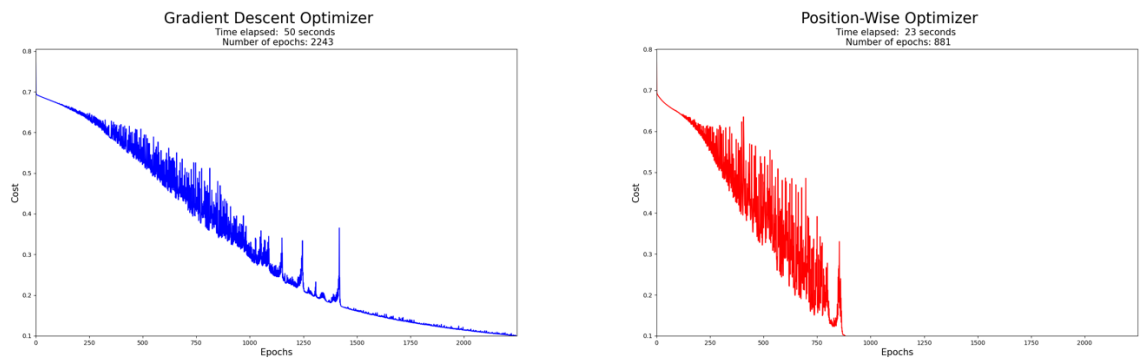

**Figure 2**; The loss of each model plotted against the number of epochs for the in-house cat classification dataset. The time elapsed to train each model to achieve the desired loss value is presented above the plots.

It took 2243 epochs for the standard gradient descent method to reach a loss value of 0.1. This process took 50 seconds. Using the position-wise optimization method only took 881 epochs to reach the same value for the loss. It took only 23 seconds. Given the significant difference in the execution time of the models (50 vs 23 seconds), the position-wise model managed to reach the desired loss significantly faster than the gradient descent model.

## Experiment 2; CIFAR-10 dataset

This experiment was a multi-class classification task (10 classes). The CIFAR-10[15] dataset consists of 50000 training images. Each image is of size: (32, 32, 3). Images were reshaped into one big vector, used for the training process. The input layer was a matrix with dimensions of (3072, 50000).
The training process results for both models are presented in **Figure 3**.

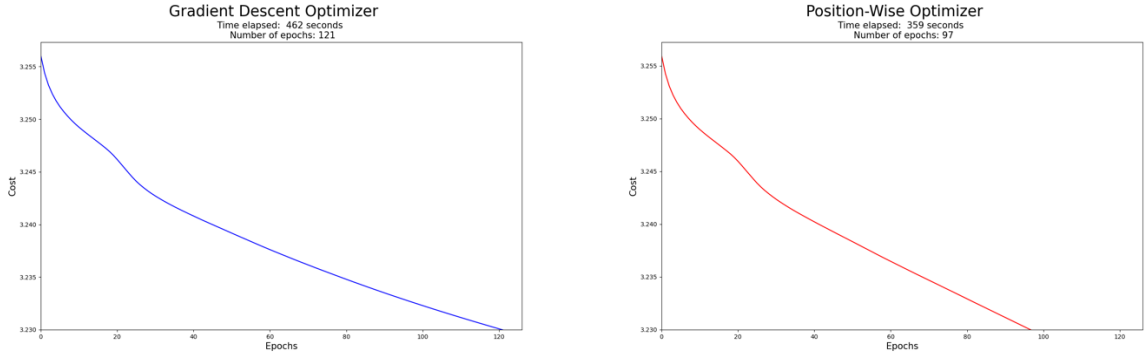

**Figure 3**; The loss of each model plotted against the number of epochs for the CIFAR-10 dataset. The time elapsed to train each model to achieve the desired loss value is presented above the plots.

It took 121 epochs for the standard gradient descent method to reach a loss value of 3.23. This process took 462 seconds. Using the position-wise optimization method only took 97 epochs to reach the same value for the loss. It took only 359 seconds. Given the significant difference in the execution time of the models (462 vs 359 seconds), the position-wise model managed to reach the desired loss significantly faster than the gradient descent model.

## Experiment 3; MNIST dataset

This experiment was a multi-class classification task (10 classes). The MNIST dataset[16] consists of 60000 training images. Each image is of size: (28, 28). Images were reshaped into one big vector, used for the training process. The input layer was a matrix with dimensions of (784, 60000).
The training process results for both models are presented in **Figure 4**.

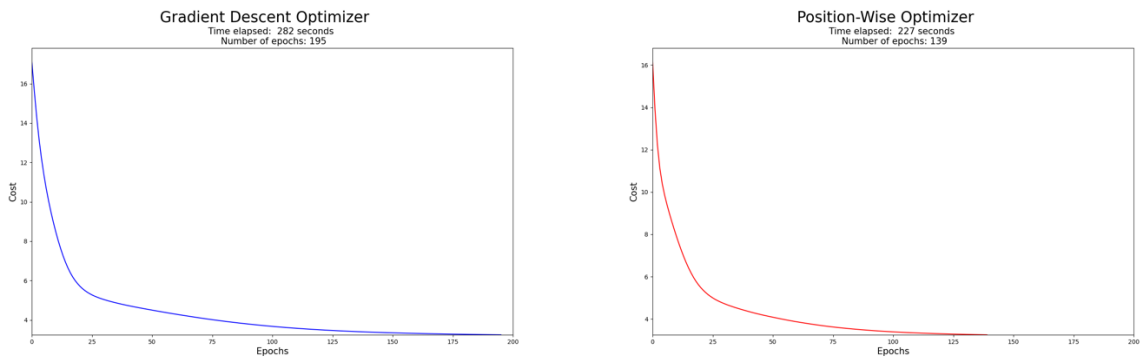

**Figure 4**; The loss of each model plotted against the number of epochs for the MNIST dataset. The time elapsed to train each model to achieve the desired loss value is presented above the plots.

It took 195 epochs for the standard gradient descent method to reach a loss value of 3.248. This process took 282 seconds. Using the position-wise optimization method only took 139 epochs to reach the same value for the loss. It took only 227 seconds. Given the significant difference in the execution time of the models (282 vs 227 seconds), the position-wise model managed to reach the desired loss significantly faster than the gradient descent model.

## Conclusions

In this manuscript, a novel nature-inspired optimization algorithm was introduced. This model imitates the pattern of synaptic plasticity of biological neural networks. The results of testing the model on three distinct datasets showed the superiority of the model over the traditional gradient descent optimization algorithm.

Synaptic plasticity in biological networks is not just about how much weights change in the synapses, but also about forming new informative synapses and terminating synapses that do not contribute to solving the optimization problem. In the model introduced in this manuscript, this phenomenon was not implemented. Future research might try to use the techniques of previous models to implement such phenomenon in the model and thus, improve upon it. It should also be noted that it is not yet proven that the human brain uses gradient methods to solve optimization problems. Thus, future researchers are recommended to try to implement synaptic plasticity behavior in other forms of optimization models as well.

# Administrative information

## Funding

No sources of support were provided for this research.

## Declarations of interest

The author declares no conflicts of interest.

## Ethics approval

Not applicable.

## Consent to participate

Not applicable.

## Consent to publication

Not applicable.

## Availability of data and material

Not applicable.

## Code availability

A simplified implementation of the model is available at the following online repository: https://github.com/thisisamirv/Position-wise-optimizer

## Author's contributions

AV was the sole author.

# References

1. von Bernhardi, R., Bernhardi, L. E. & Eugenín, J. What Is Neural Plasticity? in 1–15 (2017). doi:10.1007/978-3-319-62817-2_1.
2. Hebb, D. O. *The Organization of Behavior*. (Psychology Press, 2005). doi:10.4324/9781410612403.
3. Citri, A. & Malenka, R. C. Synaptic Plasticity: Multiple Forms, Functions, and Mechanisms. *Neuropsychopharmacology* **33**, 18–41 (2008).
4. Magee, J. C. & Grienberger, C. Synaptic Plasticity Forms and Functions. *Annual Review of Neuroscience* **43**, 95–117 (2020).
5. Geuna, S., Fornaro, M., Raimondo, S. & Giacobini-Robecchi, M. G. Plasticity and regeneration in the peripheral nervous system. *Ital J Anat Embryol* **115**, 91–4 (2010).
6. Nishizaki, S., Numao, M., Caro, J. D. L. & Suarez, M. T. C. *Theory And Practice Of Computation - Proceedings Of Workshop On Computation: Theory And Practice (Wctp2015)*. (World Scientific Publishing Company, 2017).
7. Sun, S., Cao, Z., Zhu, H. & Zhao, J. A Survey of Optimization Methods From a Machine Learning Perspective. *IEEE Transactions on Cybernetics* **50**, 3668–3681 (2020).
8. Ruder, S. An overview of gradient descent optimization algorithms. *arXiv preprint arXiv:1609.04747* (2016).
9. Lillicrap, T. P., Santoro, A., Marris, L., Akerman, C. J. & Hinton, G. Backpropagation and the brain. *Nature Reviews Neuroscience* **21**, 335–346 (2020).
10. Rumelhart, D. E., Hinton, G. E. & Williams, R. J. Learning representations by back-propagating errors. *Nature* **323**, 533–536 (1986).
11. Kaiser, J., Mostafa, H. & Neftci, E. Synaptic Plasticity Dynamics for Deep Continuous Local Learning (DECOLLE). *Frontiers in Neuroscience* **14**, (2020).
12. Negrov, D. *et al.* An approximate backpropagation learning rule for memristor based neural networks using synaptic plasticity. *Neurocomputing* **237**, 193–199 (2017).
13. Whittington, J. C. R. & Bogacz, R. An Approximation of the Error Backpropagation Algorithm in a Predictive Coding Network with Local Hebbian Synaptic Plasticity. *Neural Computation* **29**, 1229–1262 (2017).
14. Jansson, P. A. Neural Networks: An Overview. *Analytical Chemistry* **63**, 357A-362A (1991).
15. Krizhevsky, A. & Hinton, G. Learning multiple layers of features from tiny images. (2009).
16. Li Deng. The MNIST Database of Handwritten Digit Images for Machine Learning Research [Best of the Web]. *IEEE Signal Processing Magazine* **29**, 141–142 (2012).